\pdfoutput=1
\documentclass[11pt]{article}

\usepackage{acl}
\usepackage{amsmath}
\usepackage{times}
\usepackage{bm}
\usepackage{latexsym}
\usepackage{booktabs}
\usepackage{multirow}
\usepackage{amsfonts}
\usepackage{graphicx}
\newcommand\numberthis{\addtocounter{equation}{1}\tag{\theequation}}
\usepackage[T1]{fontenc}
\usepackage[utf8]{inputenc}
\usepackage{microtype}

\usepackage{inconsolata}

\title{LoRETTA: Low-Rank Economic Tensor-Train Adaptation for Ultra-Low-Parameter Fine-Tuning of Large Language Models}

\newcommand{\figref}[1]{\figurename~{\ref{#1}}}
\newcommand{\tabref}[1]{Table~{\ref{#1}}}

\newcommand{\ours}[0]{{LoRETTA} }
\newcommand{\adp}[0]{{LoRETTA}$_{adp}$ }
\newcommand{\rep}[0]{{LoRETTA}$_{rep}$ }

\author{Yifan Yang$^1$\footnotemark[1], Jiajun Zhou$^2$\footnotemark[1]\footnotemark[2], Ngai Wong$^2$ and Zheng Zhang$^1$ \\
	$^1$University of California, Santa Barbara \\
	$^2$The University of Hong Kong\\
	\texttt{yifanyang@cs.ucsb.edu}, \texttt{\{jjzhou,nwong\}@eee.hku.hk}, \texttt{zhengzhang@ece.ucsb.edu} \\
}

\begin{document}
	\maketitle
	\renewcommand{\thefootnote}{\fnsymbol{footnote}}
	\footnotetext[1]{Equal contributions}
	\footnotetext[2]{Work undertaken during the visit at UC Santa Barbara}

%%%%%%%%%%%%%%%%%%%%%%%%%%%%%%%%%%%%%%%%
%%%%%%%% -- PAPER CONTENT STARTS -- %%%%%%%%%

\begin{abstract}
	% The rapid deployment of Large Language Models (LLMs) are often hindered by significant memory and computational demands. Existing Parameter Efficient Fine-Tuning (PEFT) methods exhibit limitations due to a substantial number of trainable parameters, particularly noticeable in the recent Llama 2 models. We introduce LoRETTA, a novel framework for ultra parameter-efficient fine-tuning that dramatically reduces the number of trainable parameters through tensor-train decomposition. We propose two types of LoRETTA methods, called \adp and \rep, following the idea of adding adapters and weight reparameterization. Our implementation of \adp on the Llama 2-7b model achieves better performance compared to LoRA method with $5\times$ less parameters and achieves 90\% of accuracy with only 200K parameters. The ultra-parameter efficient LoRETTA approach successfully reduces memory requirements, diminishes the risk of over-fitting and forgetting, and enhances memory copy and computational efficiency compared to existing PEFT methods. We provide easy-to-use code built upon the Huggingface framework and PEFT library.\footnote{The code is available at \url{}}.
	
	Various parameter-efficient fine-tuning  (PEFT) techniques have been proposed to enable computationally efficient fine-tuning while maintaining model performance. However, existing PEFT methods are still limited by the growing number of trainable parameters with the rapid deployment of Large Language Models (LLMs). To address this challenge, we present LoRETTA, an ultra-parameter-efficient framework that significantly reduces trainable parameters through tensor-train decomposition. Specifically, we propose two methods, named {LoRETTA}$_{adp}$ and {LoRETTA}$_{rep}$. The former employs tensorized adapters, offering a high-performance yet lightweight approach for the fine-tuning of LLMs. The latter emphasizes fine-tuning via weight parameterization with a set of small tensor factors. LoRETTA achieves comparable or better performance than most widely used PEFT methods with up to $100\times$ fewer parameters on the LLaMA-2-7B models. Furthermore, empirical results demonstrate that the proposed method effectively improves training efficiency, enjoys better multi-task learning performance, and enhances the anti-overfitting capability. Plug-and-play codes built upon the Huggingface framework and PEFT library will be released.\footnote[3]{Code available at: \url{https://github.com/yifanycc/loretta}}

	% The fine-tuning and deployment of Large Language Models (LLMs) present challenges due to high memory requirements and computational costs. Existing approaches, such as Low-rank adaptation (~\lora) and other Parameter Efficient Fine-Tuning (PEFT) methods, aim to alleviate these challenges but face limitations concerning pre-trained LLM weights and the demand for trainable parameters. This paper introduces a memory-efficient fine-tuning approach, Tensorized ~\~\adapter (TA), leveraging tensor decomposition. TA simplifies the process by updating the tensorized matrix, enabling direct application to LLM fine-tuning. Adaptive rank decomposition reduces trainable parameters while maintaining model robustness. Empirical validations against prevalent PEFT methods, (Memory storage, GLUE tasks, Mezo Tasks and Llama, etc) The LoRETTA approach effectively reduces both memory and computation costs, mitigating the risks of overfitting and forgetting when compared to existing PEFT methods.
	
	% Openview: https://openreview.net/group?id=aclweb.org/NAACL/2024
	% DDL: Dec 15, 2023
	% including ~\lora and Prefix, underscore the efficacy of TA in significantly reducing memory overhead—a substantial advance in balancing fine-tuning efficiency and memory constraints for large language models. Empirical validation on the GLUE benchmark and application to instruction following tasks with the Llama2 7B model (<1M parameters) demonstrate the effectiveness of TA in overcoming storage challenges associated with large-scale language model adaptation. 

\end{abstract}

\section{Introduction}\label{sec:intro}

\begin{figure}
	\centering
	\includegraphics[width=0.9\linewidth]{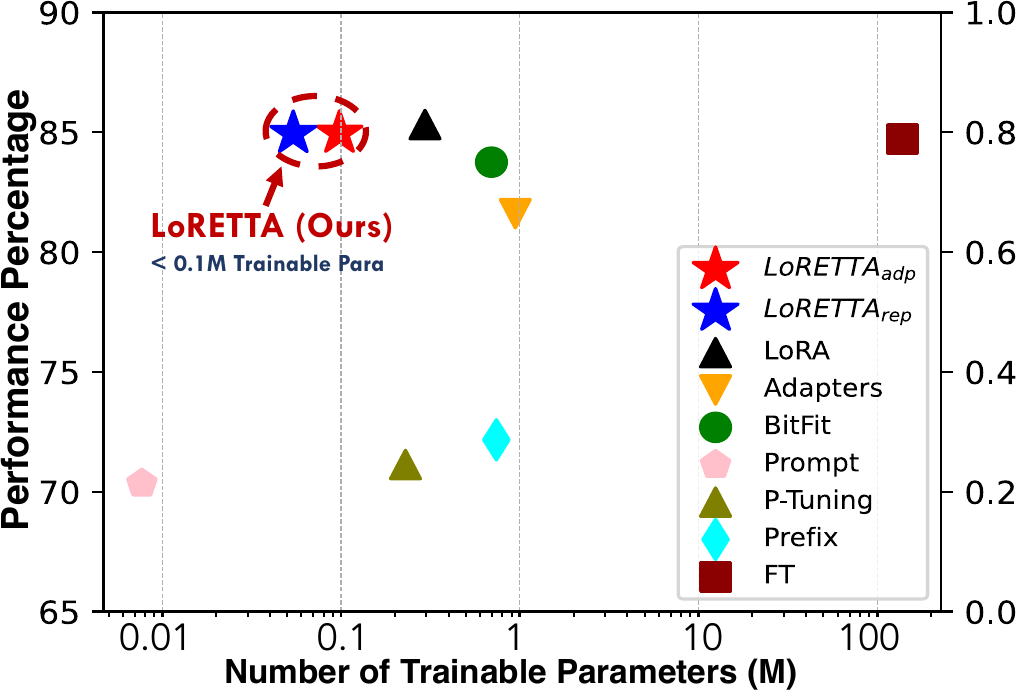}
	\caption{The performance vs. trainable parameters on the DeBERTa-Base, showcasing the relationship between parameter efficiency and performance across various GLUE tasks.}
	\label{fig:enter-label}
	\vspace{-18pt}
\end{figure}

The BERT and LLaMA families~\cite{devlin2018bert,touvron2023LLaMA,floridi2020gpt}, representing the prevailing paradigm of Large Language Models (LLMs), showcase remarkable task generalization capabilities in diverse applications, from dialogue systems to question-answering, summarization and translation. While LLMs exhibit proficiency in following instructions and learning task solutions with minimal contextual input, their accuracy can be further enhanced through fine-tuning techniques. \\~\\
Since full-model fine-tuning becomes infeasible as the model size of LLMs grows rapidly, there has been increased interest in model compression and parameter-efficient fine-tuning (PEFT)~\cite{hu2023llm, cheng2023accelerating}. PEFT methods fine-tune LLMs by modifying only a subset of parameters. The concept was initially explored in~\cite{houlsby2019parameter}, which proposes the Adapters method to inject trainable modules into the transformer encoders. Based on this concept, the LoRA approach~\cite{hu2021lora} adds low-rank updating matrices on the weights of linear projection layers in the self-attention blocks. These two types of methods achieve similar or even better performance than full-model fine-tuning, but still incur a large number of trainable parameters. Taking the LLaMA-2-70B model as an example, LoRA needs to update over 16 million parameters, which is even more than the total parameters of some BERT models. \\~\\
In contrast, other methods like prefix tuning~\cite{li2021prefix} and prompt tuning~\cite{lester2021power} introduce trainable tokens to the input or hidden layers of the base model, significantly reducing trainable parameters but potentially sacrificing accuracy, especially in few-shot learning scenarios~\cite{mao2022unipelt}. Furthermore, \cite{aghajanyan2020intrinsic} achieves approximately 90\% of the full fine-tuning performance with only 200$\sim$800 parameters on a RoBERTa model by exploring the intrinsic dimension, which is far less than the 0.3 million parameters needed in the LoRA method \cite{hu2021lora}. Despite LoRA's ability to outperform full-model fine-tuning, its number of trainable parameters is still too high, motivating our exploration of more economic and efficient high-performance PEFT approaches. This raises the question: \textit{Is there a PEFT approach with ultra-low trainable parameters that still performs on-par or better than full-model fine-tuning?}\\~\\
In this paper, we present \textbf{Lo}w-\textbf{R}ank \textbf{E}conomic \textbf{T}ensor-\textbf{T}rain \textbf{A}daptation (LoRETTA), which is tailored for efficient fine-tuning of variously scaled LLMs with minimal trainable parameters. Our approach leverages the tensor-train (TT) format to represent large weight matrices. LoRETTA encompasses two variants: \adp and \rep. The \adp variant embeds tensorized adapters in encoder/decoder layers and performs better than \emph{all} PEFT methods under equivalent trainable parameter sizes. The \rep variant, our ultra-efficient innovation, requires substantially fewer trainable parameters, occupies less than 1MB of storage, and maintains comparable performance. Our contributions are threefold:
\begin{itemize}
	\item LoRETTA is proposed that utilizes tensor-train format to effectively fine-tune LLMs with up to $100\times$ fewer trainable parameters than widely used PEFT methods like Adapters and LoRA on the LLaMA-2 model.
	% \zz{Please cite the original papers. The methods have been cited many times in intro}
	\vspace{-8.1pt}
	\item Our proposed framework demonstrates comparable performance to other widely used PEFT methods across various scales of models, tasks, and setups, particularly excelling in generation tasks with large-scale models.
	\vspace{-8.1pt}
	\item Comprehensive studies are conducted against other PEFT methods regarding storage/computation efficiency, anti-overfitting ability, forgetting risks for multi-task learning, and performance under different setups. 
	\vspace{-8.1pt}
\end{itemize}

\section{Background}

\subsection{Parameter-Efficient Fine-Tuning}\label{sec:bk-peft}
Except for the aforementioned Adapters, LoRA, and prompt-based approach, there exist various PEFT-related works~\cite{li2021prefix,lester2021power, hyeon2021fedpara, liu2023parameter}, including the BitFit method~\cite{zaken2022bitfit} that tries to further reduce trainable parameters by only fine-tuning the bias term. However, it is observed that BitFit suffers from a considerable performance drop, which is also shown in our experiments. Furthermore, there are large-scale models like LLaMA that do not employ any bias terms in the model structure, which makes the utilization of the BitFit method restricted. Compared with these previous methods, the proposed \ours is efficient and versatile, making it applicable to any kind of language model, offering a seamless and lightweight plug-and-play solution for fine-tuning. 

% Compared with these previous methods, our LoRETTA approach substantially reduces the count of trainable parameters and shows a more stable training process than the Adapters and LoRA method with much less possibility of overfitting. Our approach is versatile, making it applicable to any kind of language model, offering a seamless and lightweight plug-and-play solution for efficient fine-tuning.

\subsection{Tensor-based Model Compression}
% Several tensor decomposition methods have been developed to compress the fully connected layers, which include CANDECOMP/PARAFAC (CP) decomposition \cite{lebedev2015speeding}, Tucker decomposition \cite{kim2015compression}, and tensor-train decomposition \cite{novikov2015tensorizing}. 
% Over the past decades, tensor-compression techniques in model training have emerged as a potent solution for reducing storage requirements, and diminishing inference and training durations \cite{novikov2015tensorizing, kim2015compression}. 

Over the past decade, tensor compression has emerged as a promising technique for reducing model size and both inference and training times~\cite{lebedev2015speeding, kim2015compression}. For example,~\cite{novikov2015tensorizing} proposed the idea of the TT format by representing the weight matrix with a series of tensor factors. \citep{hawkins2022towards,hawkins2021bayesian} presented an end-to-end compressed training approach with automatic rank determination for various tensor formats. Despite these advancements, the application of the tensorized approach to the fine-tuning of LLMs is limited, primarily due to the complex, high-rank structure of pretrained weights. \\~\\
An exception to this trend is the work of~\cite{liu2021enabling}, which proposed a tensorized fine-tuning approach by only updating parts of the tensor factors. Nevertheless, it still requires over 10\% of the model parameters for effective fine-tuning. Researchers in~\cite{jie2023fact}, instead, tried to stack all weight matrices of the Vision Transformer (ViT) into a single weight tensor and create a tensorized updating tensor following the idea of LoRA. However, its applicability to LLMs is hindered by the extremely large stacked tensor, which, for the LLaMA-2-7B model, reaches 7 billion parameters for this single variable. 

% Furthermore, their methodthe reconstruction of such a large tensor is impractical on standard GPU servers, presenting a significant obstacle to its implementation in LLM contexts.
% In the realm of transformer models, which are comprised largely of linear layers, the application of tensor-train decomposition has been initially explored in convolutional neural networks (CNNs) for addressing computer vision problems \cite{novikov2015tensorizing, kim2015compression, garipov2016ultimate}. However, transformer models are potentially a better fit for tensor-based compression. In this context, \cite{ma2019tensorized} and \cite{yang2023quantization} explored using block-term tensor decomposition and tensor-train decomposition, respectively, for compressing smaller transformers models in natural language understanding tasks.

\begin{figure}[t]
	\centering
	\includegraphics[width=0.475\textwidth]{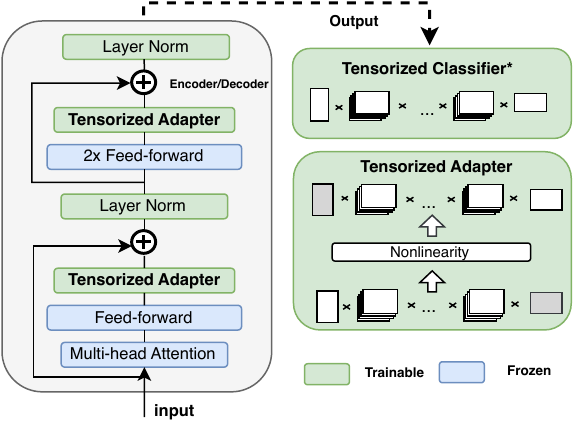}
	\caption{Architecture of \adp for the transformer encoders or decoders. $*$ the tensorized classifier is optional for different tasks. For classification tasks, we set this part to be trainable and we freeze this part during language modeling tasks.}
	\label{fig:method_1}
\end{figure}
\section{\ours Method}
PEFT methods can be broadly categorized into three types, the adapters, the reparameterization method, and the prompt-based method \cite{hu2023llm}. Among them, the reparameterization-based and adapter-based methods are notable for incorporating new structures within the model architecture, thereby introducing a large number of additional trainable parameters. To reduce the size of the injected modules, we introduce our \ours framework, which contains the adapter-based approach \adp and the reparameterization-based approach \rep. Subsequent sections will delve into the intricacies of the tensorized layer, followed by an in-depth exploration of the ~\adp and ~\rep structures.

% Our framework is characterized by several salient features:
% \begin{itemize}
	% \item Enhanced parameter efficiency, characterized by reduced latency, diminished requirement for gradient storage, and lowered risk of overfitting.
	% \item Demonstrable state-of-the-art performance with a minimal count of trainable parameters, exhibiting robust generalization capabilities across a diverse array of tasks, including but not limited to GLUE, SuperGLUE, and various generation tasks.
	% \item Versatility in application, enabling straightforward adaptation to a wide spectrum of PEFT methods.
	% \end{itemize}

\subsection{Tensorized TT Layer}\label{sec:tt-ll}
We devise the modules in \adp and \rep based on tensorized layers, where we first reshape the weight matrix in the linear layer into a tensor and then employ the TT format to reduce the number of model parameters. Specifically, TT~\cite{oseledets2011tensor} decomposes a large tensor into a set of small tensor factors. Unlike traditional linear layers that involve training large weight matrices, we only store and train the small TT factors during the fine-tuning process. Consequently, considering a fully connected layer with an input vector of $\bm{x} \in \mathbb{R}^N$, the forward pass can be expressed as $\bm{y} = \bm{W}\bm{x} + \bm{b},$ where $\bm{W} \in \mathbb{R}^{M \times N}$ is the weight matrix, and $\bm{b}$ is the bias vector. \\~\\
In a tensorized layer, the matrix $\bm{W}$ is first reshaped into a tensor $\mathcal{W} \in \mathbb{R}^{k_1 \times \cdots \times k_d}$, where $\prod_{i=1}^d k_i = M \times N$. Then, the reshaped weight tensor $\mathcal{W}$ can be effectively represented by TT-format using a set of tensor factors $\mathcal{G}_1, \cdots, \mathcal{G}_i, \cdots, \mathcal{G}_d$ with the shape of $\mathcal{G}_i \in \mathbb{R}^{r_{i-1} \times k_i \times r_i}, i\in[1, d]$. Then, for each dimension $i\in[1, d]$ and for each possible value $a_i:=1, \cdots, b_i$ in the slice of $\mathcal{W}$ at dimension $i$, the following relationship holds with a given set of TT rank $[r_0, \cdots, r_d]$:
\begin{align}
	\mathcal{W}(a_1, \cdots, a_d) 
	= \bm{G}^{a_1}_1 \cdots \bm{G}^{a_i}_i  \cdots \bm{G^{a_d}_d}
\end{align}
where $\bm{G}^{a_i}_i := \mathcal{G}_i(:, a_i, :) \in \mathbb{R}^{r_{i-1} \times r_i}$. By setting the first and last TT-ranks as $r_0 = r_d = 1$, we can obtain the value for an element in $\mathcal{W}$ by doing the matrix multiplication among the slice of each tensor factor. Since the matrices $\bm{G}^{a_i}_i$ are stacked into the tensor factor $\mathcal{G}_i$, the original weight matrix $\bm{W}$ can also be written by the TT representation, which reshapes the product of all the tensor factors:
\begin{align}\label{eq:tt-rep}
	\text{TT}(\bm{W}):= \prod_{i=1}^d \mathcal{G}_i[r_{i-1}, k_i, r_i],
\end{align}
where $\mathcal{G}_i[r_{i-1}, k_i, r_i]$ means for the $i$-th tensor factor $\mathcal{G}_i$ with the size of $r_{i-1}\times k_i \times r_i$. \\~\\
As we can see, the tensorized layer substantially reduces the parameter count for the weight matrix $\bm{W}$ from the original $M \times N$  to $\sum_{i=1}^d r_{i-1}k_ir_i$. Thus, the compression ratio is closely linked to the choice of TT ranks. For simplicity, we fix all ranks $r_i, \forall i \in [1, d-1]$ to be the constant. However, adaptive rank adjustments during training, as discussed in~\cite{hawkins2022towards}, may further enhance the performance of the \ours framework. In the following, we elaborate on how to utilize this tensorized layer in the ~\adp and ~\rep methods.

\subsection{Lightweight Tensorized Adapters}

% This section introduces our approach,called ~\adp, for ultralow-parameter fine-tuning, inspired by the concept of "intrinsic dimension" from prior research \cite{aghajanyan2020intrinsic}. This concept demonstrates that fine-tuning within a smaller parameter subspace can still be effective for learning. Despite the longstanding proposal of "intrinsic dimension," and its exploration in reparametrization methods \cite{hu2021lora}, there has been a lack of efficient compression techniques for series adapter methods. We present a novel lightweight approach that tensorize both originnal adapter blocks and classifier layers, showing superior performance compared to the current state-of-the-art (SOTA) methods with ultralow trainable parameters.
% This section introduces our tensorized adapter approach, named \adp. Our method is inspired by the concept of "intrinsic dimension" from prior research~\cite{aghajanyan2020intrinsic}. This concept demonstrates the effectiveness of fine-tuning within a smaller parameter subspace. Even though the idea of "intrinsic dimension" has been further explored in the LoRA method~\cite{hu2021lora}, its potential ability is not fully utilized. \\

\adp is inspired by the ultra-low ``intrinsic dimension'' of the language models~\cite{aghajanyan2020intrinsic}. This idea has been utilized in the previous Adapters and LoRA methods by using the bottleneck approach. However, there still exists a large gap between trainable parameters of the current PEFT methods and the "intrinsic dimension" explored in \cite{aghajanyan2020intrinsic}. This motivates us to push this idea further. In our method, we first fine-tune the LLMs by injecting tensorized adapters, demonstrating superior performance with ultra-low trainable parameters. The general workflow of \adp is illustrated in Fig.~\ref{fig:method_1}. Different from the traditional Adapters method that utilizes the bottleneck structure to reduce the trainable parameters, our tensorized adapters achieve a much better compression ratio by including two tensorized linear layers and an activation function. For example, set the hidden size of the models as 768, and the bottleneck size as 64, compared to the Adapters method with the number of trainable parameters of $2\cdot 768 \cdot 64\approx98K$ for weight matrices, \adp adds only $\sum_{i=1}^{6}(5^2\cdot 8)=1.2K$ parameters, assuming tensor shapes of $[8, 8, 8, 8, 8, 8]$ and a constant TT rank of $5$. Inspired by the idea presented in~\cite{houlsby2019parameter}, we incorporate trainable tensorized adapters following each attention and feed-forward sub-layer within the self-attention blocks. \\~\\
\textbf{Optimizable modules:} Further to fine-tuning the tensorized adapters modules, we also investigate making the layer normalization and the last layer of networks trainable. From our observations in the Appendix \ref{app:cls}, it is obvious that fine-tuning the last layer of the models is crucial for classification tasks. However, it is a common challenge to fine-tune the last layer due to its large number of parameters in models like RoBERTa and DeBERTa. To tackle this, we employ the tensorized last layer for classification tasks in our methods, thereby achieving a significant reduction in trainable parameters while maintaining effectiveness, as evidenced in our experiments. Note that we choose to freeze the last layer for language model tasks since the parameters of the language model head are inherited from the pre-trained weight.

	\begin{figure*}[t]
		\centering
		\includegraphics[width=1.0\textwidth]{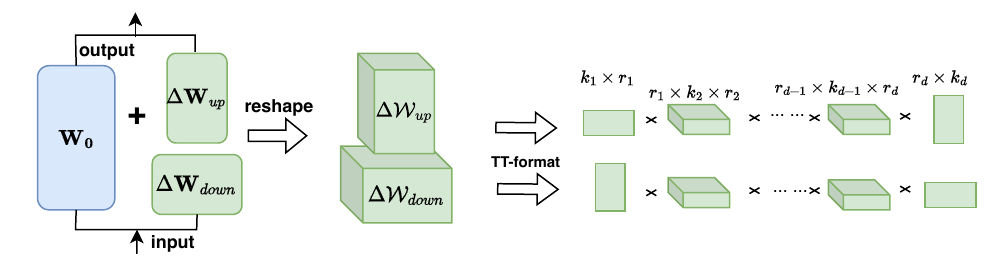}
		\caption{Architecture of the \rep method for a single transformer encoder.}
		\label{fig:method_2}
	\end{figure*}
	
	\subsection{TT Reparameterization}
	
	% ~\rep represent a reparameterization-based approach, aiming to transform weights by leveraging their low-rank subspace. This concept, initially explored in \cite{aghajanyan2020intrinsic}, demonstrates that updating a few parameters within the low intrinsic dimension subspace can preserve the capability to achieve robust performance in high-dimensional training problems of language models. Building on this, LoRA \cite{hu2021lora} introduced a novel method that decomposes the update matrix into two low-rank matrices, represented by the equation:
	Next, we propose a more compact PEFT approach by reparameterizing the weight matrix with tensor factors, dubbed ~\rep. The idea of the reparameterization also appeared in LoRA~\cite{hu2021lora}, which updates the weight with two low-rank matrices in a linear layer as follows:
	\begin{align}
		y = W_0x + \Delta Wx = W_0x + BAx
	\end{align}
	where $\bm{x}$ and $\bm{y}$ denote the input and output of a linear layer. Setting $d$ as the hidden size of the model, $W_0 \in \mathbb{R}^{d \times l}$ is a pre-trained weight matrix, $B \in \mathbb{R}^{d \times r}$ and $A \in \mathbb{R}^{r \times l}$ are low-rank matrices representing the update matrix $\Delta W$, with $r \ll$ min$(d,l)$ as the LoRA rank parameter. In the original LoRA, $A$ is initialized from a Gaussian distribution whereas $B$ is zero, ensuring that the update part $BA = 0$ at the beginning. \\~\\
	However, as mentioned in the introduction, the reparameterization of weights through matrix factorization may not fully exploit the intrinsic dimension. Here, we propose a more compact way to represent the updating matrix with two unbiased tensorized layers introduced in Section~\ref{sec:tt-ll}, whose general idea is depicted in Fig.~\ref{fig:method_2}. In our method, we also employ the bottleneck structure to first reduce the large updating matrix into two small matrices. Then, we reshape the two updating matrices $\Delta\bm{W}_{up}$ and $\Delta\bm{W}_{down}$ into tensors $\Delta\mathcal{W}_{up}$ and $\Delta\mathcal{W}_{down}$ with the shape of $k_1\times \cdots\times k_d$ and $j_1\times \cdots\times j_d$. Here, both $\Delta\mathcal{W}_{up}$ and $\Delta\mathcal{W}_{down}$ are cast into TT factors. The tensorized update process of a full-connected layer with linear transformation to an input $\bm{x}$ can be expressed as:
	% \begin{align*}
		% \bm{y} & =\bm{W}_0\bm{x} + \Delta\bm{W}_{up}\cdot \Delta\bm{W}_{down}\bm{x} \\ 
		% & = \bm{W}_0\bm{x} + \text{Reshape}(\Delta\mathcal{W}_{up}\cdot\Delta\mathcal{W}_{down})\bm{x}\\ 
		% & = \bm{W}_0\bm{x} +  \text{Reshape}(\prod_{i=1}^{d} \mathcal{G}_i\prod_{i=1}^d \mathcal{Q}_i)\bm{x} \numberthis \label{eq:updating}
		% \end{align*}\yyf{tmp, discuss tomorrow}
	\begin{align*}
		\bm{y} & =\bm{W}_0\bm{x} + \text{TT}(\Delta\bm{W}_{up})\cdot \text{TT}(\Delta\bm{W}_{down})\bm{x} \\ 
		% & = \bm{W}_0\bm{x} + \Delta\mathcal{W}_{up}^{'}\cdot\Delta\mathcal{W}_{down}^{'}\bm{x}\\ 
		& = \bm{W}_0\bm{x} +  \prod_{i=1}^{d} \mathcal{G}_i\prod_{i=1}^d \mathcal{Q}_i\bm{x} \numberthis \label{eq:updating}
	\end{align*}
	% where $\bm{W}_0$ represent the pre-trained weight and $\text{Reshape}(\cdot)$ means reshaping the result of tensor product to the shape of $\bm{W}_0$, $(\mathcal{G}_1, \cdots, \mathcal{G}_d)$ and $(\mathcal{Q}_1, \cdots, \mathcal{Q}_d)$ are tensor factors of the decomposed updating tensor $\Delta\mathcal{W}_{up}$ and $\Delta\mathcal{W}_{down}$, respectively. 
	where $\bm{W}_0$ represent the pre-trained weight, $\Delta\bm{W}_{up}$ and $\Delta\bm{W}_{down}$ are represented as the TT layers following the TT representation in eq. (\ref{eq:tt-rep}) with tensor factors $(\mathcal{G}_1, \cdots, \mathcal{G}_d)$ and $(\mathcal{Q}_1, \cdots, \mathcal{Q}_d)$ in the TT layers. In our implementation, we use the unbiased tensorized layer to perform the tensorized linear transformation in the second term of eq. (\ref{eq:updating}). In this manner, our approach reduces the parameters from 12K to 1K for a single reparameterization adapter compared with the LoRA method with the LoRA rank of 8, when the hidden size is 768 and the tensor rank is 5 for the \rep method.\\~\\
	% In this manner, our approach reduces the parameters by a factor of $\frac{d}{r}$ from $r \cdot d^2$ to $r^2 \cdot d$. Specifically, if we set the hidden size $d=768 $ and $r=4$, that implies a $192\times$ reduction).\\~\\
	\textbf{Initialization:} As noted before, LoRA starts with $B=0$, making the initial model outputs identical to pre-reparameterization. However, our proposed method requires optimizing each tensor factor. Setting a factor to zero could lead to the algorithm being stalled due to zero gradient issues. To overcome this, we initialize all tensor factors from a Gaussian distribution. Then, we assess and mitigate noise introduced by Gaussian initialization from the initialized weight matrices by conducting tensor reconstruction \cite{kolda2009tensor} during the first training step.

\section{Experiment}

We conduct comprehensive experiments for the performance of \ours on the downstream task for the LLMs with different scales. Specifically, we present the results on both BERT-family (RoBERTa-base~\cite{liu2019RoBERTa} and DeBERTa-base~\cite{he2020DeBERTa}) models and the large-scale LLaMA-2 models~\cite{touvron2023LLaMA}. We first show that \ours frameworks perform on par or better than other PEFT methods (like BitFit, LoRA, Adapters, and Prefix tuning, etc.) with fewer trainable parameters across different model types, sizes, and tasks, especially on the LLaMA-2 models. Then, we discuss some observations of the strong ability of \ours in multi-task learning and addressing overfitting issues. Further experiments demonstrate that the \ours method can help to reduce the memory storage, training FLOPs, and improve the memory copy efficiency. Finally, we also carry out the tensor rank analysis of our approach to show the applicability of \ours with even fewer trainable parameters. All experiments utilize the AdamW optimizer~\cite{loshchilov2018decoupled}, and similar learning rate and batch size set up for different methods (See Appendix \ref{app:setup} for details). We use NVIDIA Tesla V100-16GB and A100-40GB for experiments. \\~\\
% \textbf{Task Setup.} We initially conducted experiments on the Generalized Language Understanding Evaluation (GLUE) benchmark\cite{wang2018glue}, encompassing various natural language understanding tasks. These tasks include perceptual analysis (SST-2), language acceptability (CoLA), similarity and paraphrase tasks (MRPC, STS-B, QQP), and natural language reasoning (MNLI, QNLI, RTE). Subsequently, we selected both SuperGLUE tasks\cite{wang2019superglue}, involving classification (RTE, CB, BoolQ, WSC, WIC, MultiRC) and multiple-choice (COPA and ReCoRD)\yyf{check the correct corresponding to the actual experiments}, as well as two additional generation tasks about question answering (SQuAD \cite{rajpurkar2016squad}, DROP \cite{dua2019drop}). \\
\textbf{Compared Methods.} Our exploration covers both full-model fine-tuning (FT) and PEFT methods like Adapters \cite{ding2023parameter}, BitFit \cite{zaken2022bitfit}, LoRA \cite{hu2021lora}, Prefix-tuning\cite{li2021prefix}, Prompt-tuning \cite{lester2021power} and P-tuning \cite{liu2022p}. To ensure a fair and easier comparison, we implemented most PEFT methods with the Huggingface PEFT library~\cite{peft} and evaluated most methods with the same learning rate, batch size, and training epochs. Furthermore, we primarily adhered to the default settings for other hyperparameters of the baseline methods, upholding consistency across all tasks for generalizability.
\begin{table*}[t]
	\centering
	\caption{Comparative analysis of various PEFT methods on the BERT family models (including RoBERTa-base and DeBERTa-base models). The best results are bolded for the PEFT method with trainable parameters lower than 0.2M. $\ast$ represents results shown in previous works \cite{valipour2022dyLoRA, zaken2022bitfit}. Different from the LoRA paper \cite{hu2021lora}, we use the F1 score for the MRPC and QQP tasks.}
	\label{tab:glue_task}
	\resizebox{\textwidth}{!}{%
		\begin{tabular}{l|r|rrrrrrrrr}
			\hline
			Model \& Method & \begin{tabular}[c]{@{}r@{}}\# Train. \\ Param.\end{tabular} & MNLI & SST-2 & MRPC & CoLA & QNLI & QQP & RTE & STS-B & Avg. \\ \hline
			DeBERTa-Base (FT)                  & 139.19M & 88.67 & 94.61 & 91.98  & 59.32 & 93.04  &  91.42 & 68.23 & 91.10 &  84.79     \\ 
			% Bert-Large (Adapters_{r=64})                  &3.01M	&87.07	&95.041&	92.29&	60.00&	93.73&	86.81&	81.59	&90.96&	85.95     \\
			DeBERTa-Base (Adapters$_{r=8}$) & 0.94M & 87.69 & 94.72 & 88.88 & 54.19 & 92.95 & 85.52   &  59.20    &   89.68  & 81.60 \\
			DeBERTa-Base (LoRA$_{r=8}$)    & 0.30M & 87.30	&94.95	&92.84&	60.56&	93.35&	85.19&	80.14	&90.13&	85.56\\
			DeBERTa-Base (P-Tuning)    &  0.23M & 56.25	& 91.39	 &79.93 &	43.31 &	86.30&	78.43	&55.95	&78.38 & 71.24 \\
			\hline
			DeBERTa-Base (LoRA$_{r=4}$)    & 0.15M & \textbf{87.69} & 94.49 & 91.10  & 62.57 & 92.60 & 87.30  & 69.67 & 91.12 & 84.54 \\
			DeBERTa-Base (Prompt)   & 0.01M  & 77.63 &	92.43	&81.90	&32.99&	80.30	&78.15&	62.81&	56.71& 70.36 \\
			DeBERTa-Base (Prefix) &  0.15M & 60.32	&88.87	&81.22	&45.82	&83.28	& 82.22	&59.57	&84.99 & 73.28 \\
			DeBERTa-Base (BitFit)   & 0.10M &84.63	&95.41	&91.42&	\textbf{64.06}&	93.30	&84.15	&66.79&	90.23&	83.75    \\ 
			\textbf{DeBERTa-Base ($\text{LoRETTA}_{adp}$) }      & 0.10M   & 85.93 & 95.30  &\textbf{ 93.53}  & 60.84  & 92.99    & 84.08    & 75.50  & \textbf{91.32}  & \textbf{84.96} \\ 
			\textbf{DeBERTa-Base ($\text{LoRETTA}_{rep}$)  }     &  0.05M  & 86.80 & \textbf{95.53}& 88.73 & 59.69  & \textbf{93.25}  &  \textbf{89.2}   &   \textbf{75.81}  &  90.66      & 84.95 \\  
			\hline
			% RoBERTa-Base (FT)\ast   & 125M &86.4&	94.2	&\textbf{92.5}&	61.1&	92.3&	\textbf{88}&	77.4	&\textbf{90.6}&	\textbf{85.31}\\
			RoBERTa-Base (BitFit) $\ast$ & 0.1M& 85.30&	\textbf{94.80}&	\textbf{92.33}&	62.70&	91.30&	68.10	&73.60&	88.50	&82.08 \\
			RoBERTa-Base (LoRA$_{r=8}$)$\ast$   & 0.63M & \textbf{86.82}&	94.01	&91.48&	62.08	&\textbf{92.39}	&	85.71&74.51&	\textbf{90.48}& 84.69\\
			\textbf{RoBERTa-Base ($\text{LoRETTA}_{adp}$)}      & 0.10M    & 85.61&	94.38	&91.08&	\textbf{62.70}&	92.12&	\textbf{87.22}&	\textbf{78.70}&	90.26 &\textbf{85.26}\\ 
			% \textbf{RoBERTa-Base ({LoRETTA}$_{rep}$/r=5) }&  0.05M  & \\  
			\hline
		\end{tabular}
	}
\end{table*}
\subsection{GLUE Experiments on the BERT Family}\label{sec:bert}
We initially conducted experiments on the Generalized Language Understanding Evaluation (GLUE) benchmark \cite{wang2018glue}, encompassing various natural language understanding tasks. \tabref{tab:glue_task} summarizes the downstream task performance comparison between \ours framework and other baseline methods. We utilize the whole training dataset for each task, collect the best validation results in every 200 training steps, and reach the following conclusions.\\~\\
% DeBERTa enhances BERT and RoBERTa with disentangled attention and an improved mask decoder, significantly enhancing pretraining efficiency and downstream task performance—a noteworthy advancement. We utilize the pre-trained DeBERTa base (139M) from the HuggingFace Transformer library\cite{wolf2019huggingface}.\\
% \textbf{\ours demonstrates comparable or superior performance, within 1\%, across 5 out of 8 GLUE tasks on DeBERTa-Base model.} On DeBERTa-Base model, \adp has better performance with 9.5 $\times$ trainable parameters reducation in 6 GLUE task compared with Adpaters method. To fair comparision with LoRA, we take a LoRA_{r=4} case as a example, \rep reduce 1.5$\times$ parameters with around 0.8\% average performace improvement. As for other methods. like Prefix, P-tuning, Bitfir and Prompt, our methos is generality better than them.  Concerning the BERT-Large model, our \adp outperforms the Adapter method in the GLUE task with a substantial $28 \times$ reduction in trainable parameters. Results underscore our method's substantial reduction in trainable parameters by $5.4 \times$ and $9.6 \times$ compared to widely-used methods like LoRA (rank=8) and Adapter (size=8). This starkly illustrates our advantage over other prevalent Parameter Efficient Fine-Tuning (PEFT) methods across various BERT models, robustly demonstrating its generalization capabilities.
\textbf{\ours performs on-par or better than other PEFT methods.} Both \adp and \rep consistently achieve higher average scores on the GLUE tasks versus PEFT methods with lower than 0.2M trainable parameters, like LoRA, Prefix/Prompt tuning, P-tuning, and BitFit methods. Compared to LoRA with $3\times$ more trainable parameters, \adp outperforms across 4 of 8 tasks and attains a similar average performance (with nearly 0.5\% difference). Similarly, \rep reduces parameters by $6\times$ with just an average score gap within 0.6\%.\\~\\
% Notably, on the DeBERTa-Base model, \adp demonstrates an improvement of over 2\% in accuracy on the average score compared to the previous methods like BitFit with a similar parameter size. \\~\\
% \textbf{\ours achieves comparable performance to PEFT methods with significantly more trainable parameters.} Compared to LoRA with $3\times$ more trainable parameters, \ours outperforms across 4 GLUE tasks and attains a similar average score (with nearly 0.5\% difference). Similarly, \rep reduces parameters by $6\times$ with just an average score gap within 0.6\%. In comparison to \ours, other approaches like Prefix/Prompt tuning, P-tuning, and BitFit show inferior performance despite having larger training parameter sizes. \\~\\
\textbf{\ours performs well across different BERT models.} For fair comparison, we also include LoRA and BitFit results on the RoBERTa-base model reported in~\cite{valipour2022dyLoRA,zaken2022bitfit}, which sets the last layer to be trainable. We observe that \adp outperforms LoRA, with a substantial $7\times$ reduction in trainable parameters. The results also highlight \ours performs much better than the BitFit on the RoBERTa-base model, showing our advantages over other PEFT methods across various models, alongside its robust generalization capabilities.

% Please add the following required packages to your document preamble:
% \usepackage{graphicx}
\begin{table*}[ht]
	\centering
	\caption{Performance Comparison on LLaMA-2-7B with low data resource setting (1000 examples). \adp outperforms other widely used PEFT methods among most tasks.}
	\label{tab:LLaMA2_superglue}
	\resizebox{0.95\textwidth}{!}{%
		\begin{tabular}{lc|ccc|cc|cc}
			\hline
			Model \& Method &
			Train. &
			\multicolumn{3}{c|}{Classfication} &
			\multicolumn{2}{c|}{Multiple Choice} &
			\multicolumn{2}{c}{Generation} \\ \cline{3-9} 
			&
			Param.  &
			\multicolumn{1}{c|}{CB} &
			\multicolumn{1}{c|}{BoolQ} &
			\multicolumn{1}{c|}{WSC} &
			\multicolumn{1}{c|}{COPA} &
			ReCoRD &
			SQuAD &
			\multicolumn{1}{l}{DROP} \\ \hline
			LLaMA2-7B (FT)         & 6738.42M & 66.07 & 84.6 & 63.46   & 86   & \textbf{81.1} &\textbf{90.71} &51.38 \\
			LLaMA2-7B (Adapter)         & 50.33M & 66.07  & 71.8 & 62.50   & 84   & 78.8 & 88.45 & 49.14 \\
			LLaMA2-7B (LoRA$_{r=8}$)   & 4.19M & \textbf{67.86} & 84.8 & 62.50    & 81   & 79.4 & 90.56 & 45.96 \\ 
			LLaMA2-7B (Prefix)          & 1.31M  & 51.78 & 78.6 & 61.53    & 83 & 81.0 & 90.56 & 45.95 \\
			% LLaMA2-7B (LoRA_{r=2})   & 1.05M  & 69.64 & 84.9 & 62.5     & 81   & 79.5 & 90.75 & 51.6\\
			\textbf{LLaMA2-7B ({LoRETTA}$_{rep}$)} & 0.51M & 55.35 &  78.1    & 57.61  & 86   & 80.3   & 88.47 & 42.71 \\
			\textbf{LLaMA2-7B ({LoRETTA}$_{adp}$) }      & 0.88M & 66.07 & \textbf{87.0}   & \textbf{63.46}  & \textbf{87}   & 80.0   & 90.17 &\textbf{51.60} \\ \hline
		\end{tabular}%
	}
\end{table*}

\begin{table*}[ht]
	\centering
	\caption{Performance Comparison on LLaMA-2-13B and LLaMA-2-70B. We compare our proposed method with LoRA, which is one of the most widely used high-performance PEFT methods.}
	\label{tab:LLaMA2_large}
	\resizebox{0.9\textwidth}{!}{%
		\begin{tabular}{l|c|cccc|c|cc}
			\hline
			Model \& Method &
			
			\multicolumn{5}{c|}{LLaMA-2-13B} & 
			\multicolumn{3}{c}{LLaMA-2-70B} \\ 
			\cline{2-6} \cline{7-9}
			&
			Param.  &
			COPA &
			ReCoRD &
			SQuAD &
			DROP & Param.&
			SQuAD &
			DROP  \\ \hline
			LoRA$_{r=8}$              & 6.55M & 90&	83.4&\textbf{92.71}&	59.13   & 16.38M &93.78 &72.99 \\ 
			{LoRETTA}$_{rep}$ & 0.77M & 86 &  \textbf{84.4}   & 90.87 & 53.19 & 1.99M   & 90.18  &68.83   \\
			{LoRETTA}$_{adp}$     & 1.67M & \textbf{90}	& 83.9&	92.67	&\textbf{59.41}  & 4.79M   &\textbf{94.33}  & \textbf{74.50}  \\ \hline
		\end{tabular}%
	}
\end{table*}
\subsection{Large-Scale Language Models}
% Building upon the encouraging results achieved with DeBERTa, we expanded the application of \ours to the LLaMA 2 family, incorporating models at the scales of 7B and 13B (as detailed in Table \ref{tab:LLaMA2_superglue}). This extension encompasses both SuperGLUE tasks and generation tasks. For each dataset, we randomly selected 1000, 500, and 1000 examples for training, validation, and testing, respectively. The training duration for each task was set to 3 epochs, with task-specific testing for accuracy. Model checkpoints were recorded every 200 updates, and the best model was selected based on development accuracy for performance testing. Further details about the experimental settings can be found in Appendix \ref{app:setting}. The test results in Table \ref{tab:LLaMA2_superglue} led to the following observations: \\
% \subsection{Over-fitting and Multi-Task Learning}
\begin{figure}
	\centering
	\includegraphics[width=0.5\textwidth]{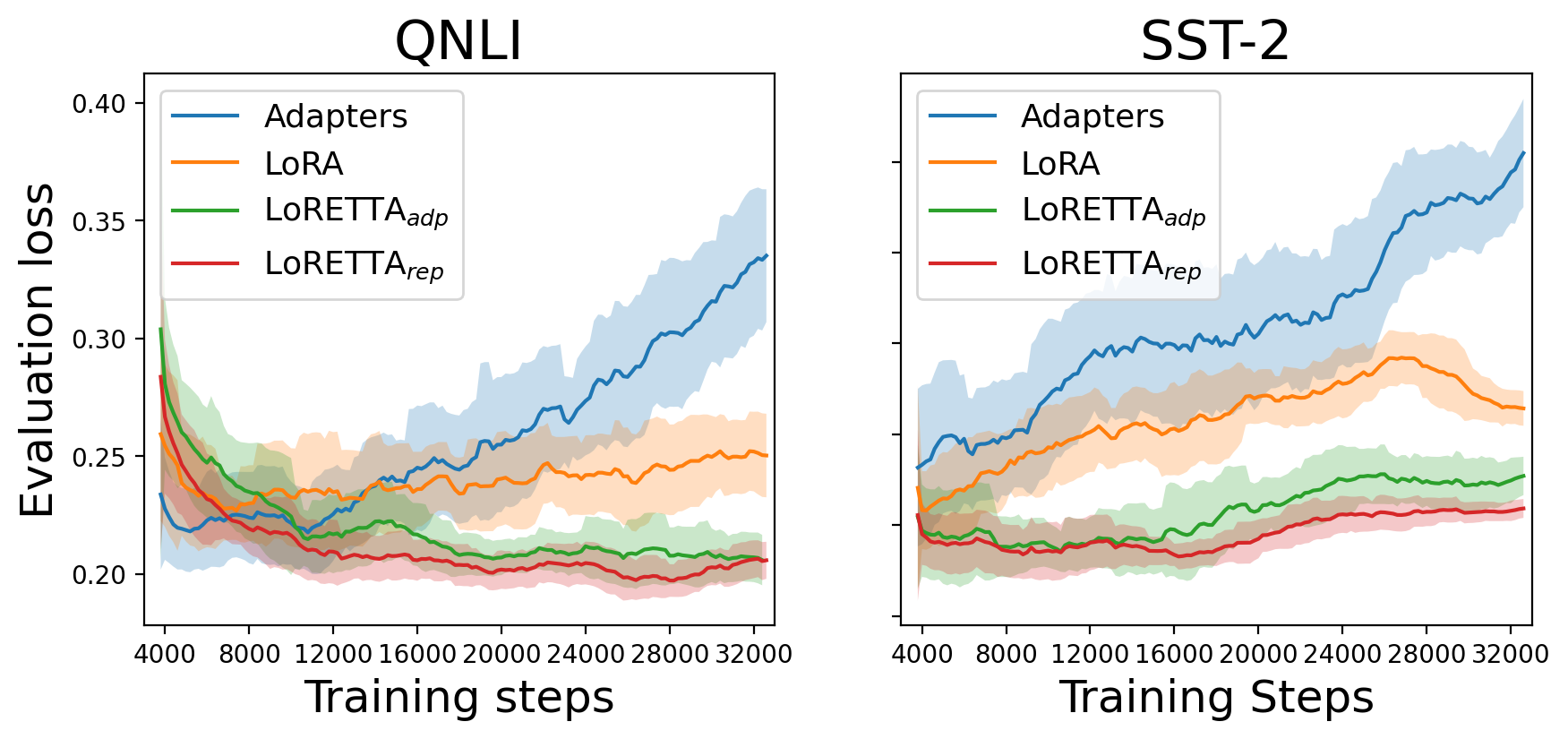}
	\caption{Evaluation loss comparison across various PEFT methods on the DeBERTa-base model. The loss is smoothed with a window size of 20 and the shallow means the standard deviation boundaries.}
	\label{fig:overfitting}
\end{figure} 
Building upon the encouraging results achieved with DeBERTa/RoBERTa models, we expanded the application of \ours to the LLaMA-2 models. The results are summarized in Table~\ref{tab:LLaMA2_superglue} and Table \ref{tab:LLaMA2_large}. To raise the difficulty of experiments, we use low data resource settings for both SuperGLUE tasks \cite{wang2019superglue} and generation tasks about question answering (SQuAD \cite{rajpurkar2016squad}, DROP \cite{dua2019drop}). For each task, we randomly selected 1000, 500, and 1000 examples for training, validation, and testing. All classification tasks in the SuperGLUE benchmark have been transferred to language modeling tasks following the prompt-based fine-tuning strategy used in \cite{malladi2023fine}. Our observations are summarized as follows. \\~\\
\textbf{\ours performs better or on-par compared with other widely used PEFT methods with up to $100\times$ trainable parameters reduction.} \adp shows superior performance across most tasks compared to \textit{all} parameter-efficient fine-tuning methods. Compared with LoRA or the Adapter methods, \adp achieves better performance in up to 7 tasks with nearly $5\times$ and $56\times$ reduction of trainable parameters. Even compared with full model fine-tuning, our method still outperforms in 5 of 7 tasks. Furthermore, \rep achieves comparable performance with up to $100\times$ fewer trainable parameters compared to the Adapters. \\~\\
\textbf{\ours is working even better on 13B and 70B models.} We compare the performance of our proposed method with the most widely used LoRA method over the LLaMA-2 13B and 70B models. Due to the limited computation resources, we only give the results on the more important reasoning (COPA and ReCoRD) and generation tasks (SQuAD and DROP). The results are summarized in Table \ref{tab:LLaMA2_large}. We can observe that our \adp method outperforms the LoRA method across 5 of 6 tasks on both 13B and 70B models. In particular, the \adp method achieves a reduction of nearly 12 million trainable parameters on the 70B model with over 1\% accuracy improvement. \\~\\
\textbf{The tensorized method shows robust performance across various tasks.} Beyond the classification and multi-choice tasks, we also included language generation tasks such as SQuAD and DROP, which are more intricate. It can be seen that \adp continues to yield excellent results with much lower trainable parameters, especially on the large-scale LLaMA-2 13B and 70B models.  

\subsection{Over-fitting and Multi-Task Learning}
\ours method uniquely addresses overfitting and promotes multi-task learning (MTL) by reducing trainable parameters. We further explore its anti-overfitting and MTL capabilities. 
\begin{table}[ht]
	\centering
	\caption{Performance of anti-forgetting in MTL tests. The three training sets are fed sequentially during the training process and we test the validation loss for each task after the training is finished.}
	\label{tab:mtl}
	\resizebox{\columnwidth}{!}{%
		\begin{tabular}{@{}lcrrr@{}}
			\toprule
			Model \& Method            & SST-2 & MRPC   & QNLI  & Average          \\ 
			\midrule
			DeBERTa-Base(Adapters)         &  51.83   &   27.21  &     90.21 & 56.42   \\
			DeBERTa-Base(LoRA)            &   49.20  &  20.15   &  87.74    & 55.70   \\
			DeBERTa-Base(\adp)       &   \textbf{52.29}  &   39.22  &    91.52 & 61.01    \\ 
			DeBERTa-Base(\rep)          &  51.26  &   \textbf{52.94}  &    \textbf{92.15}  & \textbf{65.45}    \\ 
			\bottomrule
		\end{tabular}%
	}
\end{table} ~\\
\textbf{Adapters and LoRA exhibit overfitting during training.} We follow the experiments of SST-2 and QNLI tasks in Section~\ref{sec:bert} and record the curve of evaluation loss by testing the validation dataset every 200 steps. The corresponding results are in Fig.~\ref{fig:overfitting}. It is evident from the figure that the evaluation loss for both LoRA and Adapters escalates rapidly beyond a certain point, indicating a significant over-fitting. In contrast, \adp and \rep show markedly improved handling of overfitting and a much more stable learning curve with less variance. That is attributed to their much fewer trainable parameters, which better retain the information captured by the pre-trained weights. \\~\\
\textbf{LoRETTA excels in MTL tasks.} MTL optimizes multiple tasks using shared model parameters \cite{ruder2017overview, cheng2023fc}. We utilize the DeBERTa-Base model and train our model with SST-2, MRPC, and QNLI training set in the GLUE benchmark sequentially. We test the accuracy with the validation set after the training of all three datasets, which can show the degree of forgetting. \\~\\
The results, presented in Table~\ref{tab:mtl}, demonstrate that \adp and \rep achieve higher average test accuracy. This shows our method performs better in retaining the information in the previous training, highlighting our method as a potentially better foundational approach for fine-tuning in MTL setup. Future work could include integrating more comprehensive MTL strategies with \ours, such as task clustering or task relation learning~\cite{zhang2021survey} to achieve better performance.

\subsection{Memory Performance}
\begin{figure}
	\centering
	\includegraphics[width=0.9\linewidth]{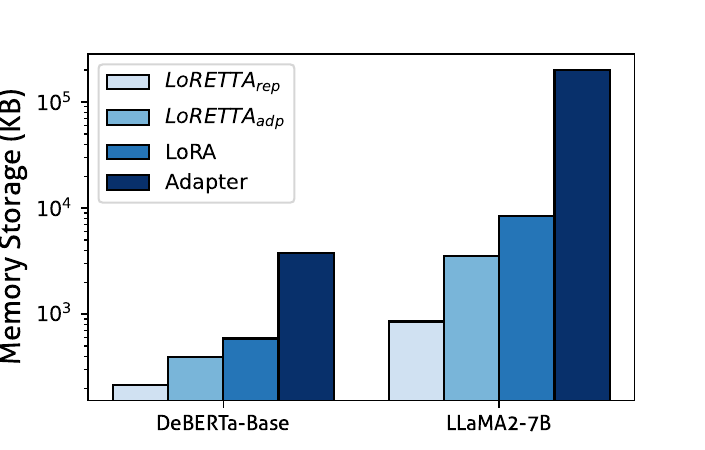}
	\caption{Comparison of memory storage for trainable parameters across different models and methods.}
	\label{fig:mem_store_train}
	\vspace{-15pt}
\end{figure}

In \figref{fig:mem_store_train}, we compare \ours with prominent fine-tuning approaches, including LoRA and adapters on two types of LLMs to show that our proposed method enjoys the following key features.\\~\\
\textbf{Ultra-low memory storage for trainable parameters.} \rep, our most compact PEFT method, requires only around 1MB storage for its trainable parameters, outperforming its counterparts. On DeBERTa-Base, both \rep and \adp (0.852MB vs 3.5MB) outperform classical baselines, reducing the trainable parameter storage by a factor of $9.6 \times$ and $2.7 \times$, respectively, compared to LoRA and Adapters. Ditto for LLaMA-2, where \rep and \adp similarly reduce the trainable parameter storage by a factor of $57.4 \times$ and $9.8 \times$, respectively. Such an economic storage space makes our proposed method suitable for resource-limited hardware~\cite{wu2023msd}, suggesting potential applications in quantized tensor models for future research.  \\~\\
\textbf{\ours minimizes data movement overhead and reduces end-to-end training FLOPs.} Considering data movement overhead during training, our method minimizes memory handling time, surpassing other PEFT methods. Overall, with $57.4 \times$ less storage consumption, \ours achieves comparable or superior results in memory copying time, as shown in \tabref{tab:flops_mem_tab}, outperforming LoRA and Adapters. It also reduces the end-to-end training FLOPs for 3 epochs in LLaMA2 fine-tuning on the SST-2 task, which shows superior computation efficiency with better accuracy. 

\begin{table}[ht]
	\centering
	\caption{Memory profiling and FLOPs analysis.}
	\label{tab:flops_mem_tab}
	\resizebox{0.48\textwidth}{!}{%
		\begin{tabular}{l|c|r}
			\hline
			Model \& Method    & Memcpy (us) & FLOPS (Reduction)   \\ \hline
			LLaMA2-7B(Adapter) & 10590       & 6.18E+15(Baseline)  \\
			LLaMA2-7B(LoRA)    & 45674       & 6.145E+15(-4.2+E13) \\
			LLaMA2-7B(\adp) & \textbf{9879} & \textbf{6.141E+15(-4.6+E13)} \\ \hline
		\end{tabular}%
	}
\end{table}

\subsection{Tensor Rank Analysis and Ablation Study}\label{sec:layer}
\begin{table}[t]
	\centering
	\caption{Tensor rank analysis on SST-2 and QNLI.}
	\label{tab:rank}
	\resizebox{0.45\textwidth}{!}{%
		\begin{tabular}{@{}l|rrrr@{}}
			\toprule
			\adp                    & r=2   & r=5    & r=10  & r=20         \\ 
			Train. Param.                     & 0.067   & 0.098    & 0.206  & 0.627 \\
			\midrule
			DeBERTa-Base(SST-2)      &   95.41   &   95.30     &  94.84   &     95.41             \\
			DeBERTa-Base(QNLI)      &   92.04   &      92.99   & 93.50     &     93.34         \\
			\midrule
			\rep                    & r=2   & r=5    & r=10  & r=20         \\ 
			Train. Param.                     & 0.042   & 0.054    & 0.094  & 0.250 \\
			\midrule
			DeBERTa-Base(SST-2)      &    94.61  &   94.4      &  94.95  &   95.07         \\
			DeBERTa-Base(QNLI)      &    92.71  &   93.25      &  93.47  &    93.32                 \\
			% \midrule
			% \adp      & r=4         & r=8           & r=16     & r=32 & r=64       \\
			% \midrule             
			%  LLaMA-2-7B(SST-2)        &      &         &    &            &             \\
			%  LLaMA-2-7B(SQuAD)       &      &         &    &            &             \\  
			%  \midrule     
			%  \rep      & r=4         & r=8           & r=16     & r=32 & r=64       \\
			% \midrule             
			%  LLaMA-2-7B(SST-2)        &      &         &    &            &             \\
			%  LLaMA-2-7B(SQuAD)       &      &         &    &            &             \\  
			\bottomrule
		\end{tabular}%
	}
	\vspace{-10pt}
\end{table}
% Finally, we analyze the structure and parameter settings for our proposed method. We compare the performance of \ours under different tensor ranks and perform an ablation to show the importance of each module we choose to optimize. \\~\\
% \textbf{Optimal rank for \ours.} 

We first investigate the influence of different tensor ranks on our model's performance. The results are summarized in Table~\ref{tab:rank}. We see that the performance for different ranks of \ours approach varies across tasks. For the SST-2 task, the performance is not sensitive to the rank setting for both \adp and \rep. However, the test accuracy drops when dealing with the QNLI task with an extra small rank. Generally, our method performs well even under smaller ranks in some tasks, which shows the possible ability to reduce the trainable parameters under tight hardware constraints. \\~\\
We also test the influence of activating the final layer and layernorm on our method. The tensorized classifier demonstrates comparable results to the regular one with a notable parameter reduction and the layernorm is shown to play a crucial role in some specific tasks. Detailed analyses are in the Appendix \ref{app:cls}.

\section{Conclusion}
% We have proposed an ultra parameter-efficient fine-tuning method, named \ours, which achieves better performance than other PEFT methods with fewer trainable parameters on LLaMA-2 models. Extensive experiments have verified that low trainable parameters can facilitate the computation and data copy demand, reduce storage requirements, and enhance the ability to deal with multi-task learning/overfitting. Our proposed methods show strong ability in both natural language understanding and generation tasks. In future work, the computation efficiency of the LoRETTA method can be further improved with other memory-efficient methods, such as FlashAttention and quantization.
We propose an ultra-parameter-efficient fine-tuning method, named \ours, which outperforms other PEFT methods with fewer trainable parameters on LLaMA-2 models. Extensive experiments have verified that having low trainable parameters can facilitate computation and memory demands, reduce storage requirements, and enhance the ability to deal with multi-task learning/overfitting. Our proposed methods exhibit strong capabilities in both natural language understanding and generation tasks. In future work, the computation efficiency of the \ours method can be further improved with other memory-efficient methods, such as FlashAttention~\cite{dao2022flashattention} and quantization~\cite{frantar2022gptq}.
\clearpage
\section*{Limitations}
Due to the numerous PEFT methods covered in this article, along with the diversity of models and tasks, the training process can be time-consuming. Exploring concepts like parallel computing in AdapterFusion~\cite{pfeiffer2020adapterfusion} could offer additional optimization possibilities. Despite our extensive experiments on LLaMA, certain modifications are still necessary. Unfortunately, the scarcity of benchmark datasets limits our ability to conduct more comprehensive experiments. In future work, there would be an extension to enhance training efficiency and scalability of the TT format, especially after adaptation to low-bit quantization~\cite{10358356}. \\~\\
 Future work will explore several directions. The ultra-small size of trainable parameters could benefit resource-limited applications. First, ~\ours can be applied for memory-efficient fine-tuning, particularly on devices with limited memory bandwidth. Second, its low parameter count is advantageous for future zeroth-order (ZO) training, where the accuracy of the ZO estimation is closely related to the dimension of the optimization problem. Third, the anti-forgetting feature of our proposed method positions it for potential use as a foundational fine-tuning approach for multi-task learning. \\~\\
Except in the field of Natural Language Processing (NLP) \cite{guo2023identifying, guo2023msq}, our proposed method can be widely adapted into other works like Automatic Speech Recognition (ASR) \cite{cheng2023ml, cheng2023ghostt5}, Vision Transformer (ViT) training \cite{xiang2022knowledge, chen2023real}, transfer learning \cite{ma2024data, ma2023transfer} and deep reinforcement learning \cite{mei2024projection}. Further experiments in these directions still need to be conducted to verify the effectiveness of the LoRETTA methods in other fields.

\section*{Ethics Statement}
\ours provides a cost-effective solution that operates with a minimal memory footprint. This alleviates the burden on data centers and reduces $CO_{2}$ emissions. However, we acknowledge that prolonged training times, especially with multiple GPUs, can pose environmental challenges. Consequently, our ongoing research endeavors are focused on developing more efficient training methods and preserving computational power with ecological considerations in mind.
% This document has been adapted
% by Steven Bethard, Ryan Cotterell and Rui Yan
% from the instructions for earlier ACL and NAACL proceedings, including those for 
% ACL 2019 by Douwe Kiela and Ivan Vuli\'{c},
% NAACL 2019 by Stephanie Lukin and Alla Roskovskaya, 
% ACL 2018 by Shay Cohen, Kevin Gimpel, and Wei Lu, 
% NAACL 2018 by Margaret Mitchell and Stephanie Lukin,
% Bib\TeX{} suggestions for (NA)ACL 2017/2018 from Jason Eisner,
% ACL 2017 by Dan Gildea and Min-Yen Kan, 
% NAACL 2017 by Margaret Mitchell, 
% ACL 2012 by Maggie Li and Michael White, 
% ACL 2010 by Jing-Shin Chang and Philipp Koehn, 
% ACL 2008 by Johanna D. Moore, Simone Teufel, James Allan, and Sadaoki Furui, 
% ACL 2005 by Hwee Tou Ng and Kemal Oflazer, 
% ACL 2002 by Eugene Charniak and Dekang Lin, 
% and earlier ACL and EACL formats written by several people, including
% John Chen, Henry S. Thompson and Donald Walker.
% Additional elements were taken from the formatting instructions of the \emph{International Joint Conference on Artificial Intelligence} and the \emph{Conference on Computer Vision and Pattern Recognition}.

% Entries for the entire Anthology, followed by custom entries
\bibliography{custom.bib}

\clearpage
\appendix
\newpage
\section{Experiment setup}\label{app:setup}
\label{sec:appendix}
\subsection{Dataset Setup} 
We initially conducted experiments on the Generalized Language Understanding Evaluation (GLUE) benchmark\cite{wang2018glue}, encompassing various natural language understanding tasks. These tasks include perceptual analysis (SST-2\cite{socher2013recursive}), language acceptability (CoLA\cite{warstadt2018neural}), similarity and paraphrase tasks (MRPC, STS-B, QQP \cite{dagan2005pascal}), and natural language reasoning (MNLI, QNLI, RTE\cite{williams2017broad, rajpurkar2018know}). The metrics we used for the GLUE benchmark are summarized in Table \ref{tab:glue_metric}.
\begin{table}[ht]
	\centering
	\caption{Metrics that we use to evaluate GLUE Benchmark for BERT-based Model.}
	\label{tab:glue_metric}
	\resizebox{0.25\textwidth}{!}{%
		\begin{tabular}{@{}cc@{}}
			\toprule
			Task Name & Metric                       \\ \midrule
			QNLI      & Accuracy                          \\
			SST-2     & Accuracy                      \\
			MNLI      & Matched Acc.          \\
			CoLA      & Matthews corr.                       \\
			MRPC      & F1                                    \\
			STS-B     & Spearman corr.                       \\
			RTE       & Accuracy                          \\
			QQP       & F1                                   \\ \bottomrule
		\end{tabular}
	}
\end{table} ~\\
Subsequently, we selected both SuperGLUE tasks~\cite{wang2019superglue}, involving classification (CB, BoolQ, WSC) and multiple-choice (COPA and ReCoRD), as well as two additional generation tasks about question answering (SQuAD~\cite{rajpurkar2016squad}, DROP~\cite{dua2019drop}). For the test with the SuperGLUE and generation datasets, we increase the difficulty by employing a low data resource setting. We randomly sample 1,000 examples for training, 500 examples for validation, and 1,000 examples for testing. We follow the prompt settings in Appendix D of~\cite{malladi2023fine} to transfer the classification into the language model tasks and the metrics we used are summarized in Table \ref{tab:superglue_metric}.
\begin{table}[ht]
	\centering
	\caption{Metrics that we use to evaluate SuperGLUE and generations tasks.}
	\label{tab:superglue_metric}
	\resizebox{0.2\textwidth}{!}{%
		\begin{tabular}{@{}cc@{}}
			\toprule
			Task Name & Metric                       \\ \midrule
			CB      & F1                          \\
			BoolQ     & Accuracy                      \\
			WSC      & F1          \\
			COPA      & Accuracy                        \\
			ReCoRD      & F1                                    \\
			SQuAD    & F1                       \\
			DROP     & F1                        \\
			\bottomrule
		\end{tabular}%
	}
	\vspace{-20pt}
\end{table}
% \subsection{Dataset}
% For DeBERTa-Base, we consider classification datasets: SST-2~\cite{socher2013recursive},  MNLI~\cite{williams2017broad}, QNLI~\cite{rajpurkar2018know}, CoLA~\cite{warstadt2018neural}, QQP, STS-B, MRPC and RTE~\cite{dagan2005pascal}. 

\subsection{Baselines}
\textbf{Fine-tuning (FT)} is a common approach for adaptation. In this process, the model is initialized with pre-trained weights and biases, and all model parameters undergo gradient updates. \\~\\
\textbf{Adapters}, as proposed by ~\cite{houlsby2019parameter}, insert adapter layers between the self-attention module (and the MLP module) and the subsequent residual connection. An adapter layer consists of two fully connected layers with biases, separated by a nonlinearity. We conducted the adapter experiment using various adapter bottleneck sizes, such as 8 and 64. \\~\\
\textbf{LoRA} introduces trainable pairs of rank decomposition matrices in parallel to existing weight matrices. As mentioned in Sections 3 and 4 \cite{hu2021lora}, we primarily apply LoRA to the query and value layers in most experiments for simplicity. The number of trainable parameters is determined by the LoRA rank and the shape of the original weights, as shown in Table~\ref{tab:hyper_para}. \\~\\
\textbf{Prefix Tuning} adds a prefix of $m$ tunable representations at each layer and freezes the remaining parts of the model. These representations serve as new keys and values, providing additional context during the attention operation. The tunable representations are initialized by randomly sampling tokens from the vocabulary and passing them through the language model to obtain their keys and values at various attention layers. In our experiments, we observe that $m$ = 8 can achieve satisfactory performance across most tasks. \\~\\
\textbf{BitFit} is a baseline where only the bias vectors are trained while keeping all other parameters frozen. We only test the BitFit methods with the BERT-based models since the bias term is not enabled in the linear layer of the LLaMA models. \\~\\
\textbf{Prompt Tuning} tuning technique can guide the behavior of language models by adding text prompts to the input, wherein we only need to train a small part of prompt parameters.

\subsection{Hyperparameters}
We outline the configuration details for each comparative experiment. Specifically, for the DeBERTa/RoBERTa-Base models, the learning rates and batch sizes of individual methods are presented in~\tabref{tab:hyper_para}. For a fair comparison, we use almost the same learning rate, batch size, and learning rate setting for different methods in the same tasks, except for the full model fine-tuning, which cannot converge under the large learning rate. In the case of P-tuning, we extended the prompt length to 768, with a virtual token count of 20 during fine-tuning. Regarding the prompt method, we increased the virtual token to 20. For prefix, we used Prefix-Propagation~\cite{li2023prefix} to experiment. We implement the LoRA, Adapters, prefix/prompt tuning, and P-tuning methods with the PEFT library \cite{peft}. All GLUE tasks underwent training for 10 to 20 epochs.\\~\\
Except for the experiments on BERT-based models, we also compare our proposed method with the Adapters, LoRA, and prefix tuning methods. We use the hyperparameters in Table \ref{tab:hyper_para_ll} for the experiment on LLaMA-2 models. Note that even though we run all experiments for 3 epochs, further learning steps may help to improve the performance of our proposed methods further. 

\subsection{Additional Detail of TT-format}
In this paper, we use the TT format to represent the weight matrices in the tensorized layer. In able to represent the weight matrices in different shapes, we design the specific shapes for models with different hidden sizes and bottleneck setups. The design of the tensor shape $[k_1, \cdots, k_d]$ is summarized in Table \ref{tab:shape}. Here we only show the tensor shape used in the DeBERTa/RoBERTa-base and LLaMA-2-7b models. The hidden sizes used are 768 and 4096 respectively. For other models with different hidden sizes, the tensor shape needs to be defined specifically before the training. More detail can be found in the code we provided, which has included the most widely used hidden sizes (like 768, 1024, 1536, 4096, 5120, and 8192) in the implementations, which work for nearly all kinds of widely used models.
\begin{table}[!ht]
	\centering
	\caption{The shape settings of the TT-format}
	\label{tab:shape}
	\resizebox{0.5\textwidth}{!}{%
		\begin{tabular}{@{}ccc@{}}
			\toprule
			Modules & Matrix Shape & Tensor Shape\\ 
			\midrule
			Tensorized Adapters         & $768\times 64$      &   [8, 8, 12, 8, 8]    \\
			& $4096\times 64$   &   [16, 16, 16, 4, 4, 4]       \\
			& $64\times 768$      &   [8, 8, 12, 8, 8]    \\
			& $64\times 4096$   &   [4, 4, 4, 16, 16, 16]       \\
			\midrule
			Tenosrized updating matrix          & $768\times 8$      &   [8, 8, 12, 8]    \\
			& $768\times 16$      &   [8, 8, 12, 4, 4]   \\
			& $768\times 32$      &   [8, 8, 12, 8, 4]   \\
			& $8\times 768$      &   [8, 12, 8, 8]    \\
			& $16\times 768$      &   [4, 4, 12, 8, 8]   \\
			& $32\times 768$      &   [4, 8, 12, 8, 8]   \\
			& $4096\times 8$   &   [8, 8, 8, 8, 8]       \\
			& $4096\times 16$   &   [8, 8, 8, 8, 4, 4]       \\
			& $4096\times 32$   &   [8, 8, 8, 8, 8, 4]       \\
			& $8\times 4096$   &   [8, 8, 8, 8, 8]       \\
			& $16\times 4096$   &   [4, 4, 8, 8, 8, 8]      \\
			& $32\times 4096$   &   [4, 8, 8, 8, 8, 8]        \\
			\midrule
			Tenosrized Classifier(Optional)         &  $768\times 768$     &    [12, 8, 8, 8, 8, 12]    \\
			&  $768\times 768$     &    [8, 8, 8, 8, 8, 8, 8, 8]    \\
			\bottomrule
		\end{tabular}%
	}
\end{table}

\section{Ablation Study on Classifier and Layernorm}\label{app:cls}
\begin{table}[ht]
	\centering
	\caption{LoRETTA fine-tuning with/without layernorm and classifier layers.}
	\label{tab:cls_layernorm_analysis}
	\resizebox{0.5\textwidth}{!}{%
		\begin{tabular}{@{}l|c|ccc|cc@{}}
			\toprule
			Method &
			\begin{tabular}[c]{@{}c@{}}Train \\ Param\end{tabular} &
			SST-2&
MRPC &
		QNLI &
\begin{tabular}[c]{@{}c@{}}Classfier\\ \& Pooler\end{tabular} &
	Layernorm \\  \midrule
			% &        &       &       &       &        Layer    &     \\ \midrule
			\adp & 0.061M & 94.38 & 92.01  & 92.98  & Tensorized & No \\
			\adp & 0.1M    & 95.3 & 95.53  & 92.99  & Tensorized & Yes  \\
			\adp & 0.650M & 93 & 91.9  & 93.15  & Regular    & No \\
			\adp & 0.688M  & 94.26 & 91.09  & 93.06  & Regular    & Yes  \\
			\adp & 0.058M  & 93.92  & 92.11  & 92.71  & No         & No \\
			\adp & 0.096M  & 94.03 &  91.31  &  93.46 & No         & Yes  \\  \midrule
			\rep & 0.054M & 95.53 &88.73& 93.25 & Tensorized & Yes \\
			\rep &  0.016M    & 93.81   &   90.78 & 90.15  & Tensorized & No \\
			\rep & 0.645M &  95.18 &  91.88  &92.99&  Regular& Yes\\
			\rep & 0.606M  &  95.41 &  91.00  &  92.57   &  Regular& No\\
			\rep & 0.052M &  95.41   &  91.19   &  92.69  & No    & Yes  \\
			\rep &  0.014M & 94.83 & 87.5 & 91.87   & No    & No \\
			\bottomrule
		\end{tabular}%
	}
\end{table}
Here, we examined six scenarios for three tasks for both \adp and \rep methods, considering the trainable status of layernorm and classifiers. The results are shown in Table \ref{tab:cls_layernorm_analysis}. Our findings highlight that the tensorized classifier demonstrates comparable results to the regular classifier with a notable reduction in parameters. Furthermore, the layernorm plays a significant role in our framework.\\~\\
First, we set the tensorized classifier/adapters to be trainable and observed the influence of layernorm. We find that layernorm plays an important role in our framework. Then, we fix the layernorm to be trainable and observe the tensorized classifier demonstrates comparable results to the regular classifier and reduces about 92\% of trainable parameters in the last layer. Furthermore, the tensorized classifier still helps a lot in improving the performance of our approach, even if we freeze the layernorm.\\~\\
We also test the influence of the tensorized classifier layer for our \rep method. As we can see from the table, optimizing the classifier layer for the sequence classification task is important. Our tensorized classifier successfully reduces the trainable parameters led by the traditional classifier layer and still maintains high performance.

\begin{table}[ht]
	\centering
	\caption{ The hyperparameter grids used for LLaMA-2 experiments. We evaluate the validation loss every 1000 steps and record the best model checkpoint according to the validation loss.}
	\label{tab:hyper_para_ll}
	\resizebox{0.45\textwidth}{!}{%
		\begin{tabular}{@{}ccc@{}}
			\toprule
			Experiment & Hyperparameters & Values \\ \midrule
			FT         & Batch size      &     $[1, 2] $    \\
			& Learning rate   &    $5e-6$        \\
			& Weight Decay    &    0    \\ \midrule
			LoRA         & Batch size      &    $[1, 2] $     \\
			& Learning rate   &     $1e-4 $    \\
			& Weight Decay    &   0     \\ 
			& Rank    &    8   \\ \midrule
			Adapters         & Batch size      &     $[1, 2]$   \\
			& Learning rate   &     $1e-4$     \\
			& Weight Decay    &    0    \\ 
			& Bottleneck $r$    &     $[8,64]$    \\ \midrule
			Prefix         & Batch size      &      $[1, 2] $   \\
			& Learning rate   &    $1e-4$     \\
			& Weight Decay    &   0    \\
			& \# Prefix Tokens    &   8     \\ \midrule
			\adp         & Batch size      &    $[1, 2]$     \\
			& Learning rate   &     $1e-4$     \\
			& Weight Decay    &     0   \\
			& Bottleneck dimension & 64 \\
			& Tensor Rank    &    $[2,4,8,16,32]$    \\\midrule
			\rep         & Batch size      &     $[1, 2] $    \\
			& Learning rate   &     $1e-4$     \\
			& Weight Decay    &   0     \\
			& Tensor Rank    &    $[2,4,8,16,32]$   \\ \bottomrule
		\end{tabular}%
	}
\end{table}

\begin{table}[ht]
	\centering
	\caption{ The hyperparameter grids used for GLUE experiments. We fine-tune each task for 10 to 20 epochs, evaluating the validation loss every 500 steps. We record the best model checkpoint based on the validation loss.}
	\label{tab:hyper_para}
	\resizebox{0.5\textwidth}{!}{%
		\begin{tabular}{@{}ccc@{}}
			\toprule
			Experiment & Hyperparameters & Values \\ \midrule
			FT         & Batch size      &    $[16, 32]$   \\
			& Learning rate   &   $1e-4$       \\
			& Weight decay    &    0    \\ \midrule
			LoRA         & Batch size      &   $[16, 32]$     \\
			& Learning rate   &    $[1e-4,5e-4]$     \\
			& Weight decay    &   0     \\ 
			& Rank    &    ${4,8}$   \\ \midrule
			Adapters         & Batch size      &    $[16, 32]$    \\
			& Learning rate   &    $[1e-4,5e-4]$     \\
			& Weight decay    &    0    \\ 
			& Bottleneck dimension    &    $[8,64]$    \\ \midrule
			Prefix         & Batch size      &     ${8,64}$   \\
			& Learning rate   &   $[1e-4,5e-4]$      \\
			& Weight decay    &   0    \\
			& \# Prefix Tokens    &   8     \\ \midrule
			Bitfit         & Batch size      &   $[16, 32]$     \\
			& Learning rate   &   $[1e-4,5e-4]$      \\
			& Weight decay    &   0   \\
			& Bias Terms    &   All    \\ \midrule
			Prompt         & Batch size      &    $[16, 32]$    \\
			& Learning rate   &    $[1e-4,5e-4]$     \\
			& Weight decay    &   0     \\
			& \# Tokens   &     10   \\ \midrule
			P-tuning         & Batch size      &    $[16, 32]$    \\
			& Learning rate   &    $[1e-4,5e-4]$     \\
			& Weight decay    &   0     \\
			& \# Tokens   &     20   \\ 
			&  Prompt Length   &     $[128,768]$  \\ \midrule
			\adp         & Batch size      &   $[16, 32]$     \\
			& Learning rate   &    $[1e-4,5e-4]$     \\
			& Weight decay    &     0   \\
			& Bottleneck dimension   &    64    \\
			& Tensor Rank    &    $[2,5,10,20]$  \\
			\midrule
			\rep         & Batch size      &    $[16, 32]$    \\
			& Learning rate   &    $[1e-4,5e-4]$     \\
			& Weight decay    &   0     \\
			& Tensor Rank    &    $[2,5,10,20]$   \\ \bottomrule
		\end{tabular}%
	}
\end{table}

%%%%%%% -- PAPER CONTENT ENDS -- %%%%%%%%

\end{document}